\def\year{2022}\relax
\documentclass[letterpaper]{article} 
\usepackage{aaai22}  
\usepackage{times}  
\usepackage{helvet}  
\usepackage{courier}  
\usepackage[hyphens]{url}  
\usepackage{graphicx} 
\usepackage{natbib}  
\usepackage{caption} 
\DeclareCaptionStyle{ruled}{labelfont=normalfont,labelsep=colon,strut=off} 
\usepackage{algorithm}
\usepackage{algorithmic}
\usepackage{arydshln}
\usepackage{amssymb}        
\usepackage{multirow}
\usepackage{verbatim}
\usepackage{amsmath}
\usepackage{booktabs}
\usepackage{xcolor}         
\usepackage{newfloat}
\usepackage{listings}
\usepackage{etoolbox}
\makeatletter
\AfterEndEnvironment{algorithm}{\let\@algcomment\relax}
\AtEndEnvironment{algorithm}{\kern2pt\hrule\relax\vskip3pt\@algcomment}
\let\@algcomment\relax
\newcommand\algcomment[1]{\def\@algcomment{\footnotesize#1}}
\renewcommand\fs@ruled{\def\@fs@cfont{\bfseries}\let\@fs@capt\floatc@ruled
  \def\@fs@pre{\hrule height.8pt depth0pt \kern2pt}%
  \def\@fs@post{}%
  \def\@fs@mid{\kern2pt\hrule\kern2pt}%
  \let\@fs@iftopcapt\iftrue}
\makeatother
\floatstyle{ruled}
\newfloat{listing}{tb}{lst}{}
\floatname{listing}{Listing}
\title{Boosting Contrastive Learning with Relation Knowledge Distillation}

\author{
    Kai Zheng,
    Yuanjiang Wang\thanks{Corresponding author(yuanjiang.wang@outlook.com).},
    Ye Yuan
}
\affiliations{
    Megvii Technology\\

    \{zhengkai, wangyuanjiang, yuanye\}@megvii.com
}

\usepackage{bibentry}

\begin{document}

\maketitle

\begin{abstract}
   While self-supervised representation learning (SSL) has proved to be effective in the large model, there is still a huge gap between the SSL and supervised method in the lightweight model when following the same solution. We delve into this problem and find that the lightweight model is prone to collapse in semantic space when simply performing instance-wise contrast. To address this issue, we propose a relation-wise contrastive paradigm with Relation Knowledge Distillation (ReKD). We introduce a heterogeneous teacher to explicitly mine the semantic information and transferring a novel relation knowledge to the student (lightweight model). The theoretical analysis supports our main concern about instance-wise contrast and verify the effectiveness of our relation-wise contrastive learning. Extensive experimental results also demonstrate that our method achieves significant improvements on multiple lightweight models. Particularly, the linear evaluation on AlexNet obviously improves the current state-of-art from $44.7\%$ to $50.1\%$ , which is the first work to get close to the supervised ($50.5\%$). Code will be made available.

\end{abstract}

\section{Introduction}\label{introduction}

\begin{figure}[ht]
\centering
\includegraphics[width=0.38\textwidth]{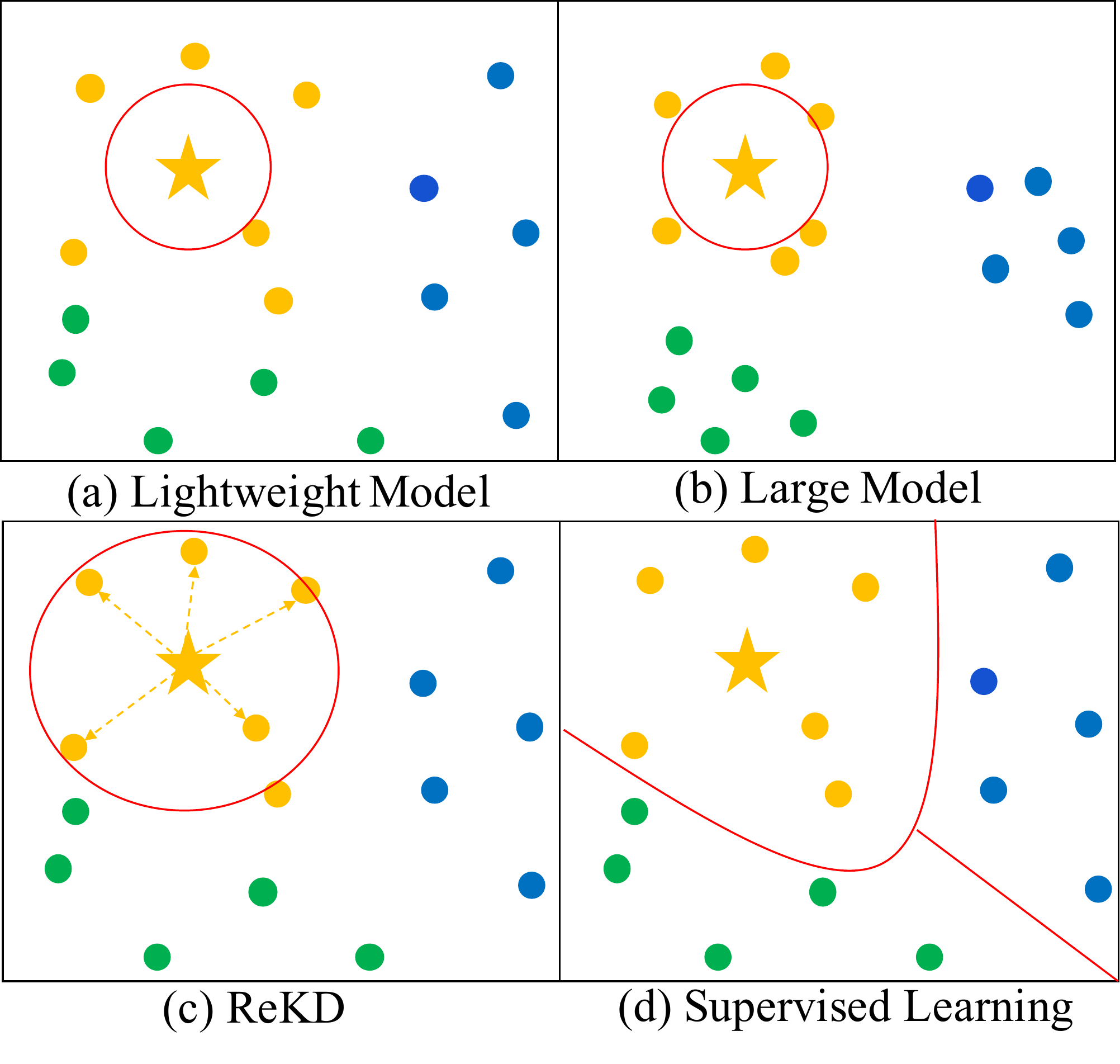}
\vspace{-3.5mm}
\caption{Different contrastive learning paradigms. Large model has better semantic feature space (a) than lightweight model (b) under the ``one-positive" instance-wise contrastive learning. (c) Our ReKD builds the relation between the instances in semantic space for lightweight model, which roles as semantic label in supervised contrastive learning (d).}
\label{fig:mot}
\vspace{-5.5mm}
\end{figure}

The rise of Deep Convolutional Neural Networks (DCNN) has led to significant success in computer vision benchmarks. Such success relies heavily on massive labeled datasets, which are prohibitively expensive to obtain. Therefore, self-supervised learning (SSL), an effective way to learn visual representations from unlabeled data, has attracted widespread attention among researchers. A variety  of different self-defined pretext tasks~\cite{komodakis2018unsupervised,feng2019self,zhang2016colorful,noroozi2016unsupervised,AET} have been proposed. Recently, instance discrimination~\cite{NPID,moco,mocov2,simclr,byol} has emerged as a dominant pretext task in unsupervised learning. This task considers each image in the dataset as an independent class, which makes the model learn discriminative feature under contrastive objective.

The renewed interest in exploring contrastive learning has derived several awesome works~\cite{mocov2, byol, simclr, swav}, some even close the gap between unsupervised-based method and supervised-based method, which attracts more researches' attention into this field. However, these works focused on how to boost the performance of large models like ResNet-50 and ResNet-50x4, and rare of them paid attention to the lightweight models like MobileNet~\cite{howard2019searching} and EfficientNet~\cite{tan2019efficientnet}. In practice, engineers prefer to select efficient models with low computational complexity and the least storage requirements, which make them easier to deploy in real-time applications, such as video surveillance and autonomous vehicles. Therefore, we pay attention to the SSL on lightweight models' performance. However, the gap between the unsupervised method and supervised method is rather huge in lightweight models. Specifically, we find that MobileNet and EfficientNet are the most typical examples, whose supervised training accuracy is 75.2\% and 77.1\%, but their unsupervised linear evaluation accuracies using MoCov2 are only 33.3\% and 37.9\%, which is far from satisfying when compared to ResNet-50's result on supervised (76.5\%) and unsupervised (67.6\%) training. Meanwhile, the similar conclusion is observed in some recent works~\cite{fang2020seed, abbasi2020compress}. Based on the practical usage of the lightweight model but with large margin exists in unsupervised training, this becomes a problem demanding prompt solution for the community.

In this work, we delve into the unsupervised learning and find that those instance-based methods all share a common issue: instances that expect to be close are undesirably pushed apart in the embedding space regardless of the intrinsic semantics in the instance, which might leads to the wrong optimization direction eventually. In our work, we term this phenomenon in SSL as \textit{semantic collapse}. We also observe that this phenomenon varies in different capacity models. The large model that has a superior feature representation, where similar semantic instances are closer than the lightweight model in the embedding space, is less likely to be involved into \textit{semantic collapse} (See Fig.~\ref{fig:mot}(a)/(b)). The tiny accuracy gap between unsupervised and supervised training in the large model proves this. In contrast, the lightweight model with low capacity may easily fall into the semantic trap if using instance-wise contrast (See Fig.~\ref{fig:mot}(a)/(d)), which is detrimental to learn a generalized feature representation.

To solve this issue, we present a \textbf{Re}lation \textbf{K}nowledge \textbf{D}istillation (ReKD) for contrastive learning, which is tailored for lightweight model with junior capacity in feature representation. In ReKD, a \textit{relation knowledge} is proposed to explicitly build the relation between the instances in the semantic space. This knowledge can alleviate the \textit{semantic collapse} existing in instance-based methods, where the semantic information inside the instance is ignored (See Fig.~\ref{fig:mot}(a)/(c)). To acquire the semantic relation knowledge for the lightweight model, we introduce a \textit{heterogeneous teacher} with a \textit{relation miner}. Given the relation knowledge, we optimize the student (lightweight model) by minimizing our proposed \textit{relation contrastive loss}.

In ReKD, we breaks ``one-positive" limitation in most instance discriminative methods~\cite{NPID,moco,simclr,byol}, and provides informative positives from the semantic level for a better contrastive objective. With this objective, the student obtains fruitful semantic knowledge from a heterogeneous teacher in the feature space, and learns a generalized representation compared with other alternatives. Furthermore, ReKD builds the bridge between clustering-based and contrastive-based method for a better self-supervised visual representation learning. Besides, ReKD is an efficient parallel computing method compared to recent self-supervised knowledge distillation (SSKD) methods~\cite{abbasi2020compress,fang2020seed} that requires a long time for pre-training an \textit{offline teacher}.

Overall, the main contributions of this work include three-fold: (i) We propose a relation knowledge distillation (ReKD) framework tailored for contrastive learning, which also builds the bridge linking cluster-based SSL and contrastive-based SSL. (ii) We provide some insights to demonstrate that our relation knowledge can help mitigate the semantic collapse theoretically. (iii) ReKD achieves a significant boost in multiple lightweight models. Notably, the SSL result on AlexNet almost close the gap with supervised learning. Meanwhile, the improvement against SSKD also verifies our method's effectiveness.

\section{Related Work}
\paragraph{Instance Discriminative learning.}
Instance discriminative based methods~\cite{oord2018representation,NPID} formulate a contrastive learning to learn feature representation, which usually contrasts between one positive and multiple negatives. MoCo~\cite{moco} and SimCLR~\cite{simclr} obtain the positive from another view generated by the data augmentation on the same image and contrast them against massive negatives. All these methods can be summarized as instance discrimination work that regards an instance as a single class. BYOL~\cite{byol} and SimSiam~\cite{simsiam} come up with a negative-free method, which achieves a competitive result only by constraining the similarity between positives without any negative instances. However, to our knowledge, these instance discrimination works all have an inescapable defect: treating all the instances as independent classes. This deficiency may lead all the instances to be separated apart regardless of whether they belong to the same semantic class or not, which can hurt the semantic-level representation in the model, especially in the lightweight model.

\paragraph{Knowledge Distillation.}
Knowledge distillation aims to transfer the knowledge learned by a larger model to a smaller one without losing important information. Many forms of knowledge and distillation strategies have been proposed to explore the best way for knowledge distillation. ~\cite{hinton2015distilling} proposes using logits with temperature to transfer the category distribution from teacher to student as additional supervision besides the original classification loss. ~\cite{romero2014fitnets,komodakis2017paying} distill the knowledge via feature/attention map. ~\cite{park2019relational} involves the mutual relation of data samples as the knowledge. ~\cite{chen2019two,park2020feature,shen2019customizing} propose the multi-teacher scheme to provide diverse knowledge from different teachers to benefit the student. Recently, some works~\cite{abbasi2020compress,fang2020seed} extend the knowledge distillation into self-supervised learning, which formulates the knowledge as the probability distribution. To our knowledge, most of these methods rely heavily on a powerful teacher model that requires a long time pre-training, namely offline teacher. This offline distillation pays little attention to the compatibility with the student and the vast time cost brought by training a teacher. In our work, we look into whether an online teacher can perform as well as an offline teacher or even better. Besides, we formulate an online relation knowledge distillation that is tailored for semantic contrastive objective.

\begin{figure*}[ht]
\centering
\includegraphics[scale=0.65]{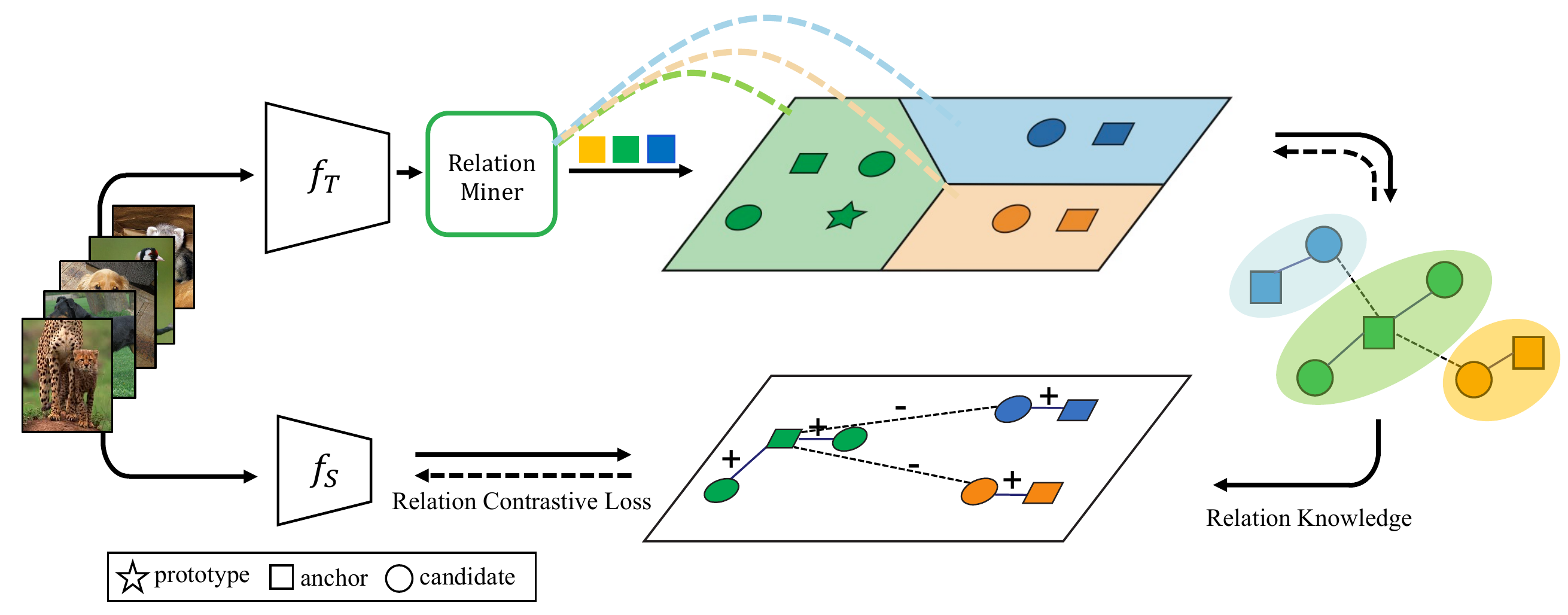}
\vspace{-1.5mm}
\caption{The pipeline of ReKD. A batch of images is fed into the heterogeneous teacher $f_T$ and student $f_S$ simultaneously. The features from heterogeneous teacher go through a relation miner, where the online clustering strategy builds the relation between the candidate and the input anchor through a bank of semantic prototypes. The relation topological structure from the relation miner serves as the relation knowledge to the student for distillation. With the relation contrastive loss, the student and heterogeneous teacher can optimize towards the semantic contrastive objective.}

\label{fig:pipeline}
\vspace{-4.5mm}
\end{figure*}

\paragraph{Clustering-based Learning.}
Early methods~\cite{xie2016unsupervised,yang2016joint,yang2017towards,chang2017deep} aim at integrating clustering into representation learning, which use the semantic clustering results as the supervision to optimize the network. DeepCluster~\cite{caron2018deep} uses the clustering labels as the pseudo label to train a classification network. LocalAgg~\cite{localagg} proposes the concept of close neighbor and background neighbor and aims to divide all the samples within the cluster around an anchor into two types. SwAV~\cite{swav} treats the label assignment as an optimal transport problem to get the clustering result and optimize the network. Apart from the conventional clustering,~\cite{deepsemantic} formulates the clustering process as optimization of network in terms of the cluster assignment constraint. Inspired by these works, we find that the cluster-based method can provide extra semantic information in terms of the clustering result. Hence, in our work, we aim to leverage clustering to benefit contrastive learning with semantic information.

\section{Preliminary}

\paragraph{Knowledge Distillation.}
Knowledge distillation~\cite{hinton2015distilling} suggests that the knowledge transferred from an influential teacher model can provide rich information for the student to learn. The objective for this task is to minimize the prediction error between the teacher and the student, which can be summarized as $\mathcal{L}_{distill}=Dist(z^T,z^S)$, where $z^T, z^S$ are the representations (e.g. softmax logit or feature) from teacher and student respectively. $Dist(\cdot)$ is the similarity metric. Although some works~\cite{abbasi2020compress, fang2020seed} have extended the distillation into self-supervised learning with response-based knowledge, it is still worth exploring whether this is the optimal knowledge for SSKD.

\paragraph{Instance Discriminative Learning.}
Instance discriminative based methods~\cite{NPID,moco,simclr} formulate an instance-wise contrastive objective to learn representation by contrasting the positive with negative. For each image $x_i$ from the training set, the encoder $f(\cdot)$ maps $x_i$ to $z_i$ with $z_i=f(x_i)$. Then the encoder is optimized by an instance-wise contrastive loss function, such as NCE~\cite{oord2018representation}. In mean teacher based methods~\cite{moco, mocov2}, $z_i^{'}=f^{'}(x_i)$ is generated as the positive from the mean teacher (a.k.a momentum encoder). Thus, we can rewrite the NCE from the distillation perspective:

\begin{equation}\small
\label{infoNCE}
\begin{aligned}
\mathcal{L}^{NCE} &= -\mathrm{log}\frac{\mathrm{exp}(z_i \cdot z_i^{'}/\tau)}{\mathrm{exp}(z_i\cdot z_i^{'}/\tau)+ \sum\limits_{n\in D(i)} \mathrm{exp}(z_i \cdot z_n/\tau)} \\
&= \mathrm{log}(1+\sum\limits_{n\in D(i)} \mathrm{exp}(z_i\cdot z_n/\tau)\cdot \mathrm{exp}(-z_i\cdot z_i^{'}/\tau))
\end{aligned}
\end{equation}

\noindent where $D(i)$ is the negative feature set for instance $i$, and $\tau$ is the temperature parameter.
In the Eq.~\ref{infoNCE}, $\mathrm{exp}(-z_i\cdot z_i^{'}/\tau)$ seeks to maximize the similarity for $z_i$ and $z_i^{'}$, which pushes the student's prediction $z_i$ close to mean teacher's \textit{historical} prediction $z_i^{'}$. For $\sum_{n\in D(i)} \mathrm{exp}(z_i\cdot z_n/\tau)$, it aims to minimize the similarity for $z_i$ and $z_n$, which separates the negative samples apart. From the distillation perspective, $\sum_{n\in D(i)} \mathrm{exp}(z_i\cdot z_n/\tau)\cdot \mathrm{exp}(-z_i\cdot z_i^{'}/\tau)$ can be regarded as the \textit{historical distillation}, which aims to make the the student mimic mean teacher's historical prediction. In this way, we can unify the objective in these mean-teacher based method as learning the response knowledge from knowledge distillation perspective.

However, the negatives in $D(i)$ are not all correct. NCE regards all the other instances as negatives, which unavoidably involves some positives belonging to the same category. Besides, we argue that only one positive from the historical version can not provide sufficient information for contrastive objective and thus limits the potential of the student.

\section{Method}

The goal of self-supervised learning is to learn rich feature representation at the semantic level. To achieve this, we propose the \textbf{Re}lation \textbf{K}nowledge \textbf{D}istillation (ReKD) in contrastive learning, as illustrated in Fig.~\ref{fig:pipeline}, which consists of an online \textit{heterogeneous teacher}, a \textit{relation knowledge} and a \textit{relation miner}. In the following subsections, we firstly introduce the \textbf{\textit{online heterogeneous teacher}} and its role in the whole framework in section Online Heterogeneous Teacher. Secondly, we customize a novel knowledge named \textbf{\textit{relation knowledge}} 
to capture the semantic information, which indicates the semantic positive/negative relationship mined by the \textbf{\textit{relation miner}} in section Relation Knowledge. Thirdly, we formulate a \textbf{\textit{relation contrastive loss}} for student's optimization in section Relation Contrastive Loss.

\subsection{Online Heterogeneous Teacher}\label{Method:Heterogeneous_Teacher}
Assumes that we have a \textit{heterogeneous teacher} module $f_T$ and a student module $f_S$. Our objective is to make the student module learn the representation from \textit{heterogeneous teacher} module to escape from the \textit{semantic collapse} phenomenon. The \textit{heterogeneous teacher} and the student use architectures from different families to ensure the superior feature representation in distillation. We maintain two candidate sets $D_t=\{u_1^t,...,u_L^t\}$ and $D_s=\{u_1^s,...,u_L^s\}$ to store the feature from the teacher $f_T$ and student $f_S$ respectively. Different from the normal self-supervised knowledge distillation methods~\cite{fang2020seed, abbasi2020compress}, we explore the online mechanism for the teacher. In our online \textit{heterogeneous teacher}, the teacher evolves simultaneously in the distillation stage, which is more parallelly efficient.

\subsection{Relation Knowledge}\label{Method:Relation Knowledge}
\begin{figure}[tb]
\centering
\includegraphics[width=8.3cm]{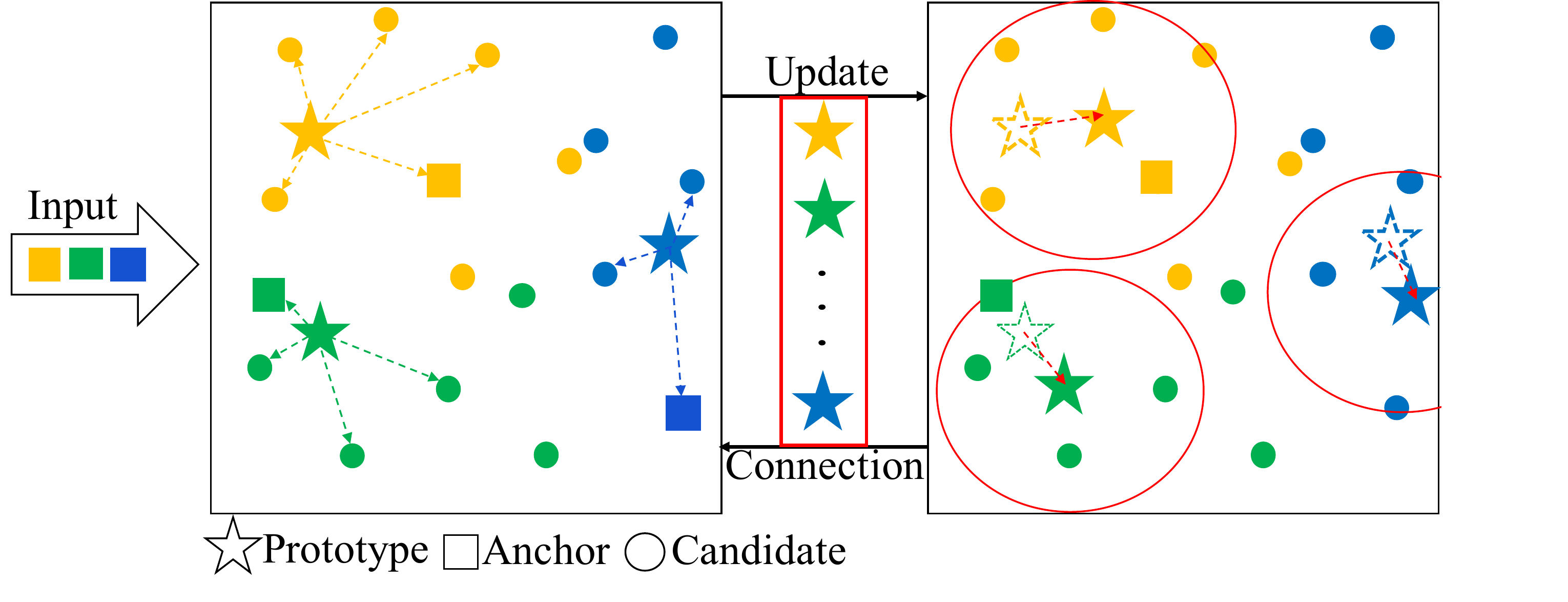}
\vspace{-2.5mm}
\caption{Illustration of Relation Miner. Given a list of input anchors, the \textit{relation miner} connects candidates with anchors by prototype bank and updates prototype bank alternately during the whole training stage.}
\label{fig:relationminer}
\vspace{-4.5mm}
\end{figure}

To alleviate the effect of semantic collapse brought by response knowledge that neglects the relation between instances and limits to historical positive, we try to explicitly model the semantic relation to introduce more diverse positives. To achieve this, we formulate a \textit{relation knowledge}, which captures the semantic positive/negative relationship for each pair of \textit{anchor} $z_i$ from the mini-batch and the \textit{candidate} $u_j$ from the candidate set $D_t$. The relation is inferred by \textit{relation miner} (see Fig.~\ref{fig:relationminer}) in the \textit{heterogeneous teacher}'s embedding space. Then, the relation is transferred to student as the guidance for contrastive objective. To achieve this, we maintain a semantic prototype bank $\mathcal{P}=\{p_1,...,p_M\}$, each prototype in the bank represents an independent semantic category learned by the model. The prototypes are initialized by spherical $k$-means clusters' centroids at the beginning of the training. After obtaining the prototype bank, the relation miner is then evolved by \textbf{connection} step and \textbf{update} step alternatively.

\paragraph{Connection.} In this step, we take the prototype in $\mathcal{P}$ as the reference to mine the relation between embeddings. Given an \textit{anchor} embedding $z_i$ and a \textit{candidate} embedding $u_j$, we can calculate the pairwise similarity for both embeddings with prototype $p_k$. For simplicity, we denote the anchor and candidate embedding uniformly as $e$, then we assign the embedding $e$ to the prototype assignment $Q(e)$ as follows:

\begin{equation}\label{protoindex}
Q(e)=
\begin{cases}
  \mathop{\arg\max}\limits_{k} \mathcal{S}(e,p_k), & \mathrm{max}\{\mathcal{S}(e,p_k)|p_k\in \mathcal{P}\}\geq \theta \\
  -1, & \mathrm{otherwise}
\end{cases}
\end{equation}

\noindent where $\mathcal{S}(e,p)$ denotes the similarity for each embedding-prototype pair. Note that if the maximum similarity is lower than $\theta$ (a threshold hyperparameter), we attribute to prototype \textnormal{\texttt{-1}}, which means this embedding fails in matching any prototypes.

After assigning the \textit{anchor} and \textit{candidate} with the corresponding prototype, we define the relation for each \textit{anchor-candidate} pair. We define it as \textit{positive} pair if the anchor and candidate have the same prototype assignment $Q$, otherwise we match them as \textit{negative} pair. Thus, for each pair, we obtain a relation (positive or negative). This relation is the core of \textit{relation knowledge} for further distillation, which contains the semantic information based on the semantic prototype retrieval. Then, we can form the positive set $P(i)$, and the negative set $N(i)$ for \textit{anchor} $z_i$, which is inferred by the relation miner. With this semantic relation, we can introduce diverse semantic positives for student, rather than limits to the low-level historical positive in the optimization.

\paragraph{Update.} In this step, we momentum updates the prototypes simultaneously rather than keep frozen or re-initial frequently. In each mini-batch, we update the feature of the \textit{anchor} $z_i$ into assigned prototype $p_k$ in terms of Eq.~\ref{update1}:

\begin{eqnarray}\label{update1}
p_k \leftarrow (1-m)z_i + mp_k & m=1-(1-\beta)\cdot \mathcal{S}(z_i,p_k)
\end{eqnarray}

\noindent where $m$ is the similarity-based coefficient controlling the weight of the \textit{anchor} embedding when updating the prototype. The range of $m$ is [$\beta$,1], where $\beta$ is usually set to a large value to ensure the prototypes to be stable and robust to unexpected noises.

\subsection{Relation Contrastive Loss}\label{Method:Relation Loss}
Based on the \textit{relation knowledge} (i.e., $P(i)$ and $N(i)$) produced by the relation miner, we propose our overall relation-wise contrast objective, namely \textit{relation contrastive loss}, which is a more generalized contrastive loss allowing for multiple positive based on relation. Unlike the historical distillation in Eq.~\ref{infoNCE}, where the student only seeks to maximize the likelihood for historical positive $u_i$ from mean teacher with respect to all negatives $u_n$ in $N(i)$, our relation contrastive loss enforce a \textit{relation} distillation to maximize the likelihood for all semantic positive $u_p$ in $P(i)$ with respect to all negatives $u_n$ in $N(i)$, which proves to be a more reasonable optimization from semantic perspective. 

\begin{equation}\label{ht}
\resizebox{1.0\linewidth}{!}
{
$\mathcal{L}^{RelCon}\!=\!
\mathrm{log}(1\!+\!\sum\limits_{n\in N(i)}\mathrm{exp}(z_i\!\cdot\! u_n/\tau)\!\cdot\! \sum\limits_{p\in P(i)}\mathrm{exp}(-z_i\!\cdot\! u_p/\tau))$
}
\end{equation}

With this semantic relation knowledge, the self-supervised model can involve the semantic information and learn more generalized representation in contrast to instance-wise contrast methods, which can solve the semantic collapse efficiently. In the following subsection, we theoretically prove why the \textit{heterogeneous teacher} and our \textit{relation knowledge} work.

\subsection{Theoretical Analysis on ReKD}\label{Discussion}

To prove how the \textit{relation knowledge} benefits in ReKD, we firstly delve into the contribution of positives and negatives in Eq.~\ref{ht}:

\begin{equation}\label{GinfoNCE}
\begin{aligned}
\mathcal{L}^{RelCon} &= \mathrm{log}(1+\sum\limits_{n\in N(i)} \mathrm{exp}(s_n)\cdot \sum\limits_{p\in P(i)}\mathrm{exp}(-s_p))
\end{aligned}
\end{equation}

\noindent where $s_p=z_i\cdot u_p/\tau$, $s_n=z_i\cdot u_n/\tau$ and $N(i),P(i)$ represent the negatives and positives set. Compared with original $\mathcal{L}^{NCE}$ in Eq.~\ref{infoNCE}, we observe that the $\mathcal{L}^{RelCon}$ involves more positives rather than only the historical one. From the positive and negative perspective, we disassemble the $N(i)=TN(i)+FN(i)$ and $P(i)=TP(i)+FP(i)$, where $TN(i),TP(i)$ denote the true negatives and true positives for \textit{anchor} $i$, $FN(i),FP(i)$ denote the false negatives and false positives. Then, we have the equation:

\begin{equation}\label{GinfoNCE2}
\resizebox{1.0\linewidth}{!}
{
\begin{math}
\begin{aligned}
\mathcal{L}^{RelCon} &= \mathrm{log}\left\{1+\frac{\sum\limits_{n\in N(i)} \mathrm{exp}(s_n)}{\sum\limits_{p\in P(i)}\mathrm{exp}(s_p)}\right\} \\
&= \mathrm{log}\left\{1+\frac{\sum\limits_{tn\in TN(i)} \mathrm{exp}(s_{tn})+\sum\limits_{fn\in FN(i)} \mathrm{exp}(s_{fn})}{\sum\limits_{tp\in TP(i)}\mathrm{exp}(s_{tp})+\sum\limits_{fp\in FP(i)}\mathrm{exp}(s_{fp})}\right\} \\
\end{aligned}
\end{math}
}
\end{equation}

Due to workable inequation of $s_{tp} > s_{tn}$ and $s_{fn} > s_{fp}$, we have the inequation:

\begin{equation}\label{GinfoNCE3}
\resizebox{1.0\linewidth}{!}
{
\begin{math}
\begin{aligned}
\sum\limits_{tp\in TP(i)} \mathrm{exp}(s_{tp})\cdot \sum\limits_{fn\in FN(i)}\mathrm{exp}(s_{fn}) > \sum\limits_{tn\in TN(i)} \mathrm{exp}(s_{tn})\cdot \sum\limits_{fp\in FP(i)}\mathrm{exp}(s_{fp})
\end{aligned}
\end{math}
}
\end{equation}

Then we apply this inequation in the Eq.~\ref{GinfoNCE2} and have the formula as follows:

\begin{equation}\label{GinfoNCE4}
\resizebox{1.0\linewidth}{!}
{
\begin{math}
\begin{aligned}
\mathcal{L}^{RelCon}
&= \mathrm{log}\left\{1+\frac{\sum\limits_{tn\in TN(i)} \mathrm{exp}(s_{tn})+\sum\limits_{fn\in FN(i)} \mathrm{exp}(s_{fn})}{\sum\limits_{tp\in TP(i)}\mathrm{exp}(s_{tp})+\sum\limits_{fp\in FP(i)}\mathrm{exp}(s_{fp})}\right\} \\
&> \mathrm{log}\left\{1+\frac{\sum\limits_{tn\in TN(i)} \mathrm{exp}(s_{tn})}{\sum\limits_{tp\in TP(i)}\mathrm{exp}(s_{tp})}\right\}
\end{aligned}
\end{math}
}
\end{equation}

from the Eq.~\ref{GinfoNCE4}, the low bound of $\mathcal{L}^{RelCon}$ can be treated as the $N(i),P(i)$ with pure true negatives $TN(i)$ and true positives $TP(i)$, which indicates that the NCE with incorrect negatives in $N(i)$ can be harmful in optimization. Besides, we conjecture that the purity of positives $TP(i)/P(i)$ and the true positive number $|TP(i)|$ is the point of contrastive based method. 

We conclude that the more accurate the \textit{relation knowledge} is, the better performance the student can achieve. This also explains the phenomenon that the supervised-based training surpasses the unsupervised one by a large margin, especially in the lightweight model. With this in mind, our ReKD narrows the gap between supervised and unsupervised for student efficiently, which also mitigates the semantic collapse. Furthermore, the experiments in section \textit{Performance of Relation Knowledge} (in Appendix) support this theoretical analysis.

\section{Experiments}\label{Exp}
In this section, we demonstrate the performance of ReKD by a standard linear evaluation protocol compared with mainstream SSL and SSKD methods.

\subsection{Representation Training Setting}
In experiments, we validate our algorithm on multiple backbones: AlexNet, MobileNet-V3, ShuffleNet-V2, EfficientNet-b0 and ResNet-18. To enable a fair comparison, we replace the last classifier layer with an MLP layer (two linear layers and one ReLU layer). The dimension of the last linear layer sets to 128. For efficient clustering, we adopt the GPU $k$-means implementation in faiss~\cite{faiss}. $M$ sets to $1000$ as default to model the dataset’s semantic distribution (ablation of $M$ refers to appendix).

\subsection{Representation Evaluation with Self-supervised Method}
To evaluate the representation, we freeze the features of the encoder of the self-supervised pre-trained model and train a single classifier layer (a fully connected layer followed by a \textit{softmax}). All the hyperparameters of linear evaluation are strictly kept aligned with the implementations in ~\cite{mocov2}.

To compare with other self-supervised methods on lightweight model, we report the accuracy of AlexNet on ImageNet in Tab.\ref{table:alexnet}, where ReKD use ResNet-50 as teacher and AlexNet as student. It is worthy to note that ReKD achieves a significant $7.2\%$ improvement than MoCov2 baseline, which is the first work comparable with the supervised learning. The improvement implies that our ReKD obviously mitigates semantic collapse in lightweight model and that the relation knowledge may role as semantic label in supervised contrastive learning.

To demonstrate that the ReKD can be generalized on different light backbones with types of large models(heterogeneous teachers). We use AlexNet, MobileNet-V3, ShuffleNet-V2, EfficientNet-b0 and ResNet-18 as lightweight models, and use ResNet-50 and ResNet-101 as large model. Tab.\ref{table:stu-tea} shows that all lightweight models achieve a consistent and significant improvements, where EfficientNet-b0 increases almost $25\%$ top-1 accuracy. This result validates that ReKD is an effective method, that can also be implemented with various lightweight models and large models (heterogeneous teachers) in a flexible way.

\begin{table}[t]
\vspace{-4.5mm}
\renewcommand\arraystretch{1.15} 
\centering
\scalebox{0.78}{
\begin{tabular}{lccccc}

\toprule
\bf{Method} & \bf{Conv1} & \bf{Conv2} & \bf{Conv3} & \bf{Conv4} & \bf{Conv5} \\ \hline
\textbf{Supervised} & 19.3 & 36.3 & 44.2 & 48.3 & 50.5 \\ \midrule
DeepCluster & 12.9 & 29.2 & 38.2 & 39.8 & 36.1 \\
NPID & 16.8 & 26.5 & 31.8 & 34.1 & 35.6 \\
LA & 14.9 & 30.1 & 35.7 & 39.4 & 40.2 \\
ODC & \textbf{19.6} & 32.8 & 40.4 & 41.4 & 37.3 \\
Rot-Decouple & 19.3 & \textbf{33.3} & 40.8 & 41.8 & 44.3 \\
MoCov2 $\dagger$ & 17.4 & 27.7 & 38.1 & 40.6 & 42.9 \\ 
SimCLR $\dagger$ & 6.0 & 30.9 & 37.7 & 42.0 & 40.3 \\ 
BYOL $\dagger$ & 7.4 & 32.0 & 39.7 & 43.9 & 44.7 \\
SwAV $\dagger$ & 11.4 & 29.4 & 34.5 & 40.4 & 44.2\\
SimSiam $\dagger$ & 17.3 & 26.8 & 37.6 & 39.8 & 44.5 \\ \midrule
\bf{ReKD (ours)} & 16.3 & 33.0 & \textbf{42.9}\small{\textcolor[RGB]{10,123,58}{(+3.2)}} & \textbf{48.1}\small{\textcolor[RGB]{10,123,58}{(+4.2)}} & \textbf{50.1}\small{\textcolor[RGB]{10,123,58}{(+5.4)}} \\
\bottomrule
\end{tabular}
}
\caption{ImageNet test accuracy (\%) using linear classification on different self-supervised learning methods. $\dagger$ denotes the result reproduced by us on AlexNet backbone.}
\label{table:alexnet}
\vspace{-5.5mm}
\end{table}

\begin{table}[h]
\centering

\scalebox{0.85}{
\begin{tabular}{lccccc}
\toprule
\multirow{2}*{\shortstack{\bf{Method}}} & \multirow{2}*{\shortstack{\bf{Student}}} & \multicolumn{3}{c}{\bf{ImageNet Acc.}} \\
\cmidrule(r){3-5}
 & & Top-1 & Top-5 \\ \midrule
\textbf{Supervised} & AlexNet & 50.5 & -- \\ \midrule
CC & AlexNet & 37.3 & -- \\
SEED $\ast$ & AlexNet & 44.7 & 69.0 \\ 
CompRess $\dagger$ & AlexNet & 46.8 & 71.3 \\ 
\bf{ReKD (ours)} & AlexNet & \textbf{50.1}\small{\textcolor[RGB]{10,123,58}{(+3.3)}} & \textbf{74.4}\small{\textcolor[RGB]{10,123,58}{(+3.1)}} \\ \midrule
SEED & R-18 & 57.6 & 81.8 \\ 
\bf{ReKD (ours)} & R-18 & \textbf{59.6}\small{\textcolor[RGB]{10,123,58}{(+2.0)}} & \textbf{83.3}\small{\textcolor[RGB]{10,123,58}{(+1.5)}} \\ \midrule
SEED & Mob-v3 & 55.2 & 80.3 \\ 
\bf{ReKD (ours)} & Mob-v3 & \textbf{56.7}\small{\textcolor[RGB]{10,123,58}{(+1.5)}} & \textbf{81.2}\small{\textcolor[RGB]{10,123,58}{(+0.9)}} \\ \midrule
SEED & Eff-b0 & 61.3 & 82.7 \\ 
\bf{ReKD (ours)} & Eff-b0 & \textbf{63.4}\small{\textcolor[RGB]{10,123,58}{(+2.1)}} & \textbf{84.3}\small{\textcolor[RGB]{10,123,58}{(+1.6)}} \\
\bottomrule

\end{tabular}}
\caption{ImageNet test accuracy (\%) using linear classification on different self-supervised knowledge distillation methods. $\ast$ indicates the result is reproduced by us. $\dagger$ denotes the result reproduced in the same architecture for fair comparison.}
\label{table:sskd}
\vspace{-5.5mm}
\end{table}

\begin{table*}[ht]
\renewcommand\arraystretch{1.2} 
\centering
\begin{tabular}{cccccccccccc}
\toprule
\multirow{2}*{\shortstack{\bf{Method}}} &  & \multicolumn{2}{c}{\bf{Alex}} & \multicolumn{2}{c}{\bf{Mob-v3}} & \multicolumn{2}{c}{\bf{Shuff-v2}} & \multicolumn{2}{c}{\bf{Eff-b0}} & \multicolumn{2}{c}{\bf{R-18}}\\
 & \textbf{Teacher} & T-1 & T-5 & T-1 & T-5 & T-1 & T-5 & T-1 & T-5 & T-1 & T-5 \\ \midrule
\bf{Supervised} & -- & \multicolumn{2}{c}{50.5} & \multicolumn{2}{c}{75.2} & \multicolumn{2}{c}{75.4} & \multicolumn{2}{c}{77.5} & \multicolumn{2}{c}{69.8}\\ \hdashline
\bf{Self-supervised} & \multirow{2}*{\shortstack{--}} & \multirow{2}*{\shortstack{42.9}} & \multirow{2}*{\shortstack{66.4}} & \multirow{2}*{\shortstack{35.3}} & \multirow{2}*{\shortstack{61.0}} & \multirow{2}*{\shortstack{52.0}} & \multirow{2}*{\shortstack{75.8}} & \multirow{2}*{\shortstack{38.6}} & \multirow{2}*{\shortstack{65.3}} & \multirow{2}*{\shortstack{53.3}} & \multirow{2}*{\shortstack{78.4}} \\ 
MoCov2  \\ \hdashline
\multirow{2}*{\shortstack{\textbf{ReKD (ours)}}} & \multirow{2}*{\shortstack{R-50 (67.6)}} & 50.1 & 74.4 & 56.7 & 81.2 & 61.9 & 83.8 & 63.4 & 84.3 & 59.6 & 83.3 \\
       & & \small{\textcolor[RGB]{10,123,58}{+7.2}} & \small{\textcolor[RGB]{10,123,58}{+8.0}} & \small{\textcolor[RGB]{10,123,58}{+21.4}} & \small{\textcolor[RGB]{10,123,58}{+20.2}} & \small{\textcolor[RGB]{10,123,58}{+9.9}} & \small{\textcolor[RGB]{10,123,58}{+8.0}} & \small{\textcolor[RGB]{10,123,58}{+24.8}} &
       \small{\textcolor[RGB]{10,123,58}{+19.0}} & \small{\textcolor[RGB]{10,123,58}{+6.3}} & \small{\textcolor[RGB]{10,123,58}{+4.9}} \\ \hdashline
\multirow{2}*{\shortstack{\textbf{ReKD (ours)}}} & \multirow{2}*{\shortstack{R-101 (69.7)}} & 50.8 & 75.1 & 59.6 & 83.1 & 63.6 & 84.9 & 65.0 & 85.7 & 59.7 & 83.9 \\
       & & \small{\textcolor[RGB]{10,123,58}{+7.9}} & \small{\textcolor[RGB]{10,123,58}{+8.7}} & \small{\textcolor[RGB]{10,123,58}{+24.3}} & \small{\textcolor[RGB]{10,123,58}{+22.1}} & \small{\textcolor[RGB]{10,123,58}{+11.6}} & \small{\textcolor[RGB]{10,123,58}{+9.1}} & \small{\textcolor[RGB]{10,123,58}{+26.4}} &
       \small{\textcolor[RGB]{10,123,58}{+20.4}} & \small{\textcolor[RGB]{10,123,58}{+6.4}} & \small{\textcolor[RGB]{10,123,58}{+5.5}} \\
\bottomrule
\end{tabular}
\caption{ImageNet test accuracy (\%) using linear classification on multiple student architectures. T-1 and T-5 denote Top-1 and Top-5 accuracy using linear evaluation. First column denotes the different methods using for training. Second column indicates Top-1 accuracy of teacher networks in MoCov2 self-supervised learning. First row indicates the student networks, while second row shows the supervised performances of student networks. Third row denotes the self-supervised baseline with MoCov2. Note, all the methods are trained for 200 epochs.}
\label{table:stu-tea}
\end{table*}

\subsection{Representation Evaluation with Self-supervised Knowledge Distillation Method}
To prove the effectiveness of ReKD, we conduct experiments with all the offline SSKD methods~\cite{fang2020seed,abbasi2020compress,noroozi2018boosting} on the same backbone (AlexNet) following the linear classification. Note that the original teacher used in~\cite{abbasi2020compress} is MoCov2 with ResNet-50 pre-trained 800 epochs offline. For a fair comparison, we change the teacher of MoCov2 with ResNet-50 of 200 epochs pre-trained model to compare with ReKD. In Tab.\ref{table:sskd}, our ReKD outperforms all the SSKD methods on AlexNet.

\section{Further Analysis}\label{Analysis}
In this section, we analyze ReKD from different perspectives.

\subsection{Ablation for Components.}
In Tab.\ref{table:ablation:teacher}, we report the impact of applying \textit{heterogeneous teacher} on the selected method~\cite{fang2020seed} (Tab.\ref{table:ablation:teacher}.b) and our method (Tab.\ref{table:ablation:teacher}.c). The baseline (Tab.\ref{table:ablation:teacher}.a) is MoCov2~\cite{mocov2} using a mean teacher to guide the student. We see that an extra \textit{heterogeneous teacher} can boost the performance by a significant margin of $2.3\%$ on Top-1 Acc. With the \textit{relation knowledge} we propose, the performance can be further improved by $7.2\%$. This validates our ReKD with relation knowledge can break the semantic collapse, which also help learn a generalized representation. 

\subsection{Response Knowledge vs. Relation Knowledge.}
To prove the effectiveness of our \textit{relation knowledge}, we compare it with a response knowledge-based method~\cite{fang2020seed} under the same offline teacher setting. We see that in Tab.\ref{table:ablation:online}, the method using the \textit{relation knowledge} with an offline teacher can improve $2.4\%$ points compared to the response knowledge with an offline teacher. The improvement suggests that the \textit{relation knowledge} would be a better choice when considering knowledge in SSKD. This also implies that the student may benefit from numerous and accurate semantic positive samples connected by the \textit{relation knowledge}, which is ignored by response knowledge~\cite{fang2020seed, abbasi2020compress}. This result also supports the theoretical analysis in the previous section \textit{How Relation Knowledge benefits}.

\begin{table}[h]

\centering
\scalebox{0.85}{
\begin{tabular}{ccccc}

\toprule
\multirow{2}*{\shortstack{\textbf{Method}}} & Heterogeneous & Relation & \multicolumn{2}{c}{\bf{ImageNet Acc.}} \\ 
\cmidrule(r){4-5}
 & Teacher & Knowledge & Top-1 & Top-5 \\
\midrule
a & & & 42.9 & 66.4 \\
b & $\surd$ & & 45.2 & 69.1 \\ \midrule
c & $\surd$ & $\surd$ & \textbf{50.1}\small{\textcolor[RGB]{10,123,58}{(+4.9)}} & \textbf{74.4}\small{\textcolor[RGB]{10,123,58}{(+5.3)}} \\
\bottomrule
\end{tabular}}
\caption{Ablation of the important components in ReKD: heterogeneous teacher and relation knowledge. The accuracy of Top-1 and Top-5 is evaluated by AlexNet for linear classification on ImageNet.}
\label{table:ablation:teacher}
\end{table}

\begin{table}[h]
\vspace{-0.5mm}
\centering
\scalebox{0.95}{
\begin{tabular}{lllccc}

\toprule
\multicolumn{3}{c}{\bf{Distillation Strategy}} & \multicolumn{3}{c}{\bf{ImageNet Acc.}} \\
\cmidrule(r){1-3} \cmidrule(r){4-6}
\multicolumn{1}{r}{Knowledge} & & Teacher & Top-1 & & Top-5 \\ \midrule
\multirow{2}*{\shortstack{response}} & & offline & 44.7 & & 69.0        \\
\multicolumn{1}{l}{}         & & online & 45.2 & & 69.1 \\ \midrule
\multirow{2}*{\shortstack{relation}} & & offline & 47.1 & & 71.3 \\
\multicolumn{1}{l}{}         & & online & \textbf{50.1} & & \textbf{74.4}   \\
\bottomrule
\end{tabular}}
\caption{Ablation of distillation strategies for knowledge type and teacher mechanism. The result is evaluated in AlexNet architecture for linear classification on ImageNet.}
\label{table:ablation:online}
\vspace{-4.5mm}
\end{table}

\subsection{Online Teacher vs. Offline Teacher.}
We do an ablation study of the online/offline teacher on ReKD. As for the offline teacher case, we first train the teacher with $200$ epoch and freeze all the trainable parameters during distillation. For the online teacher case, both teacher and student update simultaneously. In Tab.\ref{table:ablation:online}, we observe that online teacher cases outperform all the offline teacher cases, which suggests the potential of online teacher in SSKD. The same conclusion is observed in Deep Mutual Learning~\cite{zhang2018deep}. Intuitively, under the offline paradigm, the huge performance gap between teacher and student may lead to instability and low convergence for the student. Our online teacher can alleviate this and provide a proper curriculum for the student. Besides, this online teacher mechanism can save much time compared with normal offline SSKD methods, since the offline teacher consumes an extra pre-trained time. Compared with SSL methods, ReKD achieves much more improvement if allowing the equivalent extra time cost increase. For example, BYOL~\cite{byol} costs almost $100\%$ extra time due to the symmetric structure while only gains $+1.8\%$ accuracy improvement against MoCov2 in Tab.\ref{table:alexnet}. In contrast, ReKD achieves a significant $+7.2\%$ increase on accuracy while only takes a $+118\%$ extra time cost increase.

\subsection{Why Heterogeneous Teacher has better semantic representation.} 
\begin{figure}[tb]
\centering
\includegraphics[width=6cm]{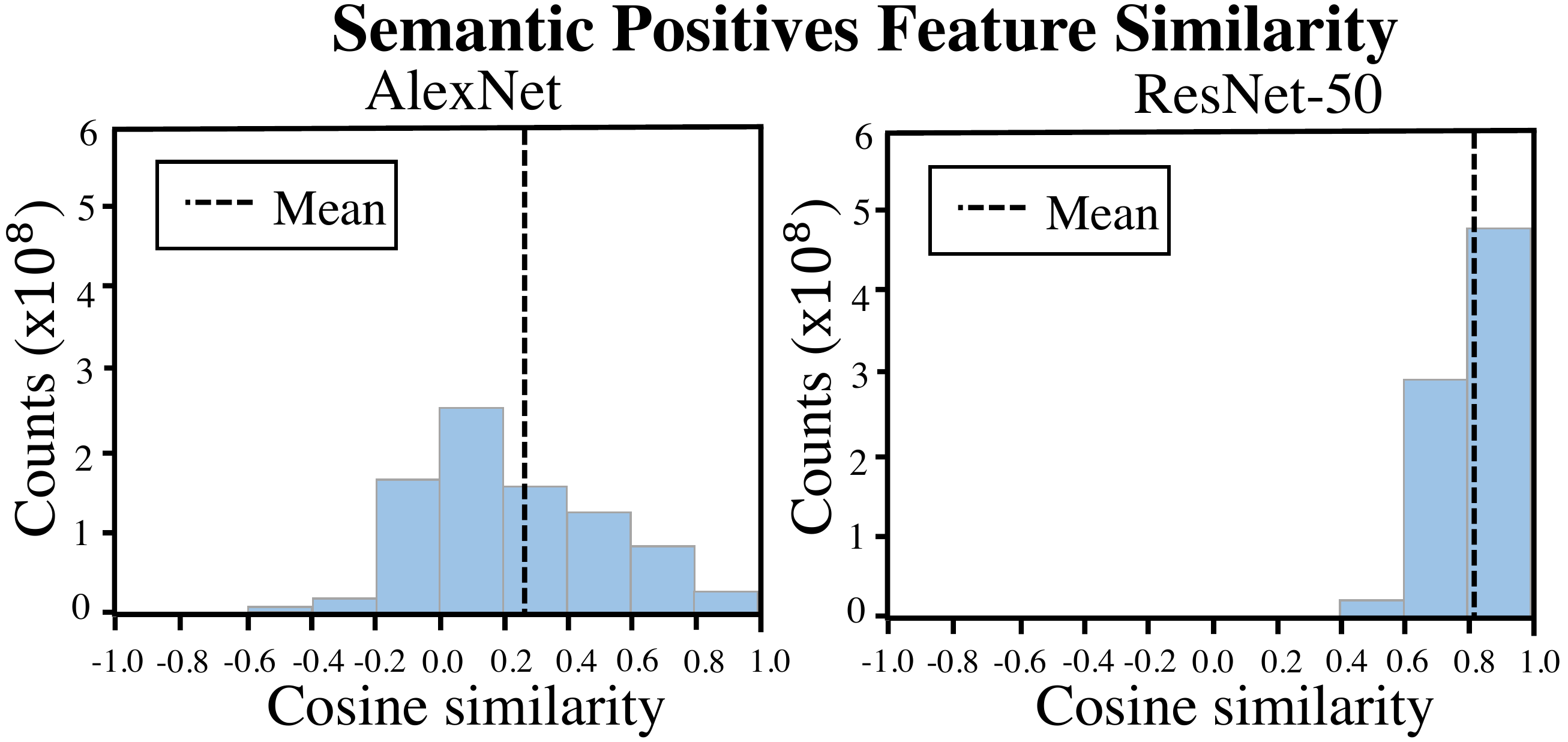}
\caption{Distribution of similarity between semantic similar pairs from different capacity of backbones.}
\label{fig:sim_distribution}
\vspace{-4.5mm}
\end{figure}

In our relation distillation, it is critically important to choose a well-performed teacher. We analyze the feature representation on the different capacity model, such as AlexNet and ResNet-50. We select MoCov2 as our unsupervised feature extractor and extract all the features with different backbones (AlexNet and ResNet-50) from images in ImageNet, then we measure the feature distance (i.e. cosine similarity) for each pair of instance with same semantic ground-truth label. Fig.~\ref{fig:sim_distribution} summarizes the resulting similarity distribution, where the features from the large model (ResNet-50) exhibit large similarity, which indicates that the large model can capture more semantic information inherently. Also, the \textnormal{\texttt{Mean}} in the figure refers to the mean similarity value of all the pairs, which indicates the overall feature extraction capacity. This result also supports the motivation (in Fig.~\ref{fig:mot}(a)/(b)) that the semantic collapse is more severe in the lightweight model under unsupervised training. Therefore, Heterogeneous teacher has better semantic representation.

\section{Conclusion}
We propose a Relation Knowledge Distillation (ReKD) to alleviate the semantic collapse in most instances discriminative-based methods. Specifically, the ReKD benefits from relation knowledge, which provides the semantic relation to guide the lightweight model for the semantic contrastive objective. The theoretical analysis supports our main concern about instance-wise contrast and verifys the effectiveness of our relation-wise contrastive learning. Our extensive experiments on SSL and SSKD benchmarks demonstrate the effectiveness of ReKD. Furthermore, we hope our work can raise the community's attention to explore the efficient distillation way for the lightweight model in self-supervised learning.

\bibliography{aaai22}

\appendix
\clearpage

\section{Training Details for Unsupervised Representation Learning}\label{appendix:train_details}
For different student backbones, we all use SGD as the optimizer. The SGD weight decay is 0.0001 and the momentum is 0.9. For AlexNet, we train with a mini-batch size of 512 on 8 RTX 2080Ti GPUs and an initial learning rate of 0.12. The temperature in the loss function sets to 0.2. For MobileNet-V3, ShuffleNet-V2 and EfficientNet-b0, we only change the min-batch size to 256, learning rate to 0.06, temperature to 0.1. For ResNet-18, the mini-batch size is 256 and the learning rate is 0.03. For all the student models, the cosine decay schedule is used to adjust the learning rate for the whole 200 epochs training phase, and the momentum parameter of the mean teacher is 0.999. In all teacher models (e.g. ResNet-50 and ResNet-101), we keep the mini-batch size as 256 with learning rate 0.03, and other hyperparameters are the same with MoCov2. The capacity of the candidate set is 65536. For the specific components in ReKD, we set the capacity of the prototype bank to 1000. The confident threshold is 0.8 and the momentum coefficient for the prototypes updating is 0.9. For more details about the ablation study for these hyper-parameters refer to the following sections.

\section{Training Details for Evaluation}\label{appendix:lincls_detail}
For linear classification, we train 100 epochs for different student encoders with the SGD optimizer. The weight decay of SGD sets to 1e-7, 0. and 0. for Alexnet, MobileNet-V3/ShuffleNet-V2/EfficientNet-b0 and ResNet-18 respectively. The initial learning rate is set to 0.01 and reduced by a linear decay for AlexNet (similar in DeepCluster). For MobileNet-V3, ShuffleNet-V2, EfficientNet-b0 and ResNet-18, we set 3 as the initial learning rate, and reduce a factor of 10 at 60 and 80 epochs (similar in MoCov2).

\section{Experiments Result on Downstream Tasks}\label{appendix:downstream}
In order to evaluate the transferrable feature learned by our model, we conduct two specific downstream tasks, object detection and segmentation. We choose Faster RCNN for object detection trained on VOC-07+12 train+val set and evaluated on VOC-07 test set. We also train Mask RCNN for both object detection and segmentation on COCO 2017 dataset. Following the same training setting as previous work, we initialize the weight of the backbone from the model that is pretrained from self-supervised method, and finetune all the layers of the model. For VOC dataset, we train the model for 48k iterations with a mini-batch size of 8. The learning rate is initialized with 0.001 and the weight decay sets to 0.0001. The image scale is [480, 800] pixels during training and 800 at inference. We use $\mathrm{AP}_{50}$, $\mathrm{AP}$, $\mathrm{AP}_{75}$ as evaluation metric on VOC-07 test set. For COCO dataset, we train MaskRCNN for 180k iterations and the initial learning rate is 0.02 with a decayed schedule at 120k and 160k iterations. The image scale is [600, 800] pixels during training and 800 at inference. We also use $\mathrm{AP}_{50}$, $\mathrm{AP}$, $\mathrm{AP}_{75}$ as the evaluation metric.

\begin{table}[h]
\centering
\resizebox{1.0\linewidth}{!}
{
\begin{tabular}{cllccc}
\toprule
\textbf{Task} & \textbf{Pre-train} & \textbf{Dataset} & $\mathrm{AP}_{50}$ & $\mathrm{AP}$ & $\mathrm{AP}_{75}$ \\ \midrule
& scratch & VOC & 65.8 & 38.1 & 38.0 \\
& supervised & VOC & 72.4 & 43.7 & 45.5 \\
& MoCov2 & VOC & 73.9 & 45.9 & 48.9 \\ \cmidrule(r){2-6}
Detection & \textbf{ReKD(ours)} & VOC & \textbf{76.3}\small{\textcolor[RGB]{10,123,58}{(+2.4)}} & \textbf{48.1}\small{\textcolor[RGB]{10,123,58}{(+2.2)}} & \textbf{51.0}\small{\textcolor[RGB]{10,123,58}{(+2.1)}} \\ \cline{2-6}
& scratch & COCO & 48.4 & 30.3 & 32.1 \\
& supervised & COCO & 50.8 & 31.9 & 33.5 \\ 
& MoCov2 & COCO & 49.4 & 31.0 & 33.0 \\ \cmidrule(r){2-6}
& \textbf{ReKD(ours)} & COCO & \textbf{51.4}\small{\textcolor[RGB]{10,123,58}{(+2.0)}} & \textbf{32.3}\small{\textcolor[RGB]{10,123,58}{(+1.3)}} & \textbf{34.3}\small{\textcolor[RGB]{10,123,58}{(+1.3)}} \\ \hline
& scratch & COCO & 45.6 & 27.3 & 28.9 \\
& supervised & COCO & 47.9 & 28.5 & 29.9 \\
Segmentation & MoCov2 & COCO & 46.4 & 27.7 & 29.0 \\ \cmidrule(r){2-6} 
& \textbf{ReKD(ours)} & COCO & \textbf{48.5}\small{\textcolor[RGB]{10,123,58}{(+2.1)}} & \textbf{29.0}\small{\textcolor[RGB]{10,123,58}{(+1.3)}} & \textbf{30.6}\small{\textcolor[RGB]{10,123,58}{(+1.6)}} \\ \bottomrule
\end{tabular}
}
\vspace{-1.5mm}
\caption{Object detection and instance segmentation fine-tuned on COCO and PASCAL VOC with ResNet-18 backbone.}
\label{table:coco-finetune}
\vspace{-4.5mm}
\end{table}

Tab.\ref{table:coco-finetune} shows that ReKD not only outperforms its supervised learning counterpart by $4.4\%$ on VOC dataset in terms of $\mathrm{AP}$ in the detection task, but also surpasses MoCov2 baseline by $2.4\%$. In the segmentation task, ReKD also shows its leading performance compared with the supervised or MoCov2 pre-trained model. This result demonstrates the transferability of the unsupervisedly learned representation by our ReKD.

\section{Ablation for Hyperparameters}\label{appendix:ablationofhyperparameter}
We study the effects of different hyper-parameters in our methods using AlexNet (student) and ResNet-50 (teacher). We list the Top-1 test accuracy on ImageNet using linear evaluation protocol. With the increasing of parameter $\beta$ (a hyperparameter in computing momentum value updating to the prototype bank), ReKD gets the peak accuracy at $0.8$, as shown in Tab.\ref{table:momentum}. For the number of prototypes in the prototype bank, we do the ablation and find that $M=1000$ is the best value (See in Tab.\ref{table:kmeans}). For the similarity threshold $\theta$ in assigning prototype, we choose $0.8$ as our default since the result is insensitive to this value.

\begin{table}[htb]
\centering
\resizebox{1.0\linewidth}{!}
{
\begin{tabular}{lcccccc}
\toprule
Param. $\beta$ & 0.6 & 0.7 & 0.8 & 0.9 & 0.99 & 0.999 \\ \midrule
Top-1 Acc. & 49.2 & 49.4 & \textbf{50.1} & 49.6 & 49.5 & 49.5 \\ \bottomrule
\end{tabular}
}
\vspace{-1.2mm}
\caption{Linear evaluation accuracy (\%) of different $\beta$ in momentum value.}
\label{table:momentum}
\vspace{-4.2mm}
\end{table}

\begin{table}[htb]
\centering
\resizebox{1.0\linewidth}{!}
{
\begin{tabular}{lcccccc}
\toprule
Param. $M$ & 100 & 500 & 1000 & 2000 & 5000 & 10000 \\ \midrule
Top-1 Acc. & 47.7 & 49.5 & \textbf{50.1} & 49.4 & 49.2 & 49.5\\ \bottomrule
\end{tabular}
}
\vspace{-1.2mm}
\caption{Linear evaluation accuracy (\%) of different $M$ in prototype bank.}
\label{table:kmeans}
\vspace{-4.2mm}
\end{table}

\begin{figure*}[t]
\centering
\includegraphics[scale=0.51]{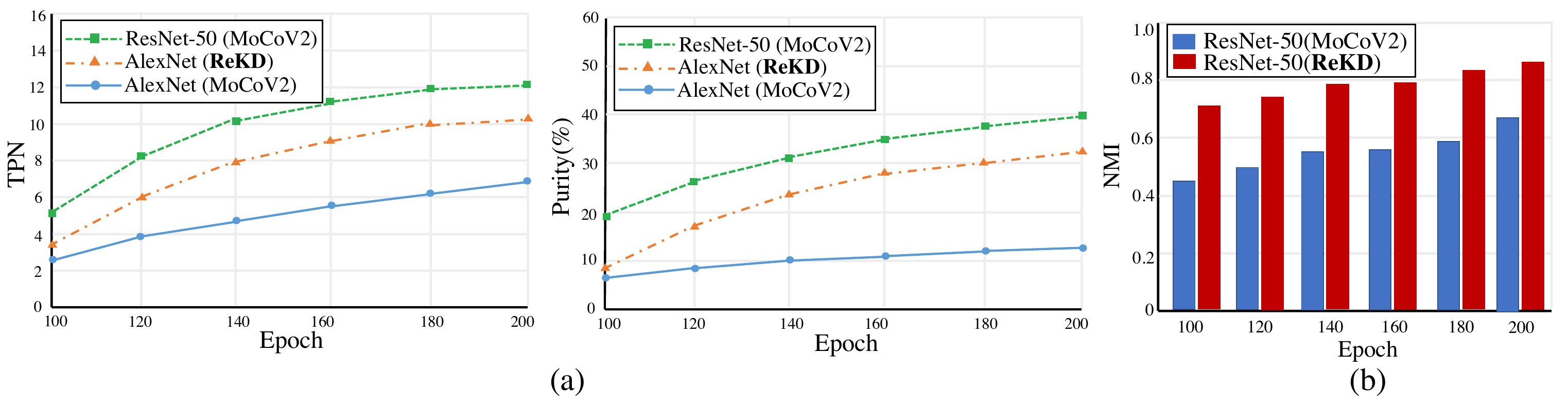}
\vspace{-2.5mm}
\caption{(a) \textnormal{\texttt{True Positive Number}} (\textnormal{\texttt{TPN}}) and \textnormal{\texttt{Purity}} calculated on MoCov2 and ReKD. Using ReKD can yield higher \textnormal{\texttt{TPN}} and \textnormal{\texttt{Purity}} than MoCov2 under the same backbone (AlexNet). (b) Normalized Mutual Information (NMI) between assigned prototype and class label.}
\label{fig:tpn_purity}
\vspace{-1.5mm}
\end{figure*}

\begin{figure*}[t]
\centering
\includegraphics[width=14cm]{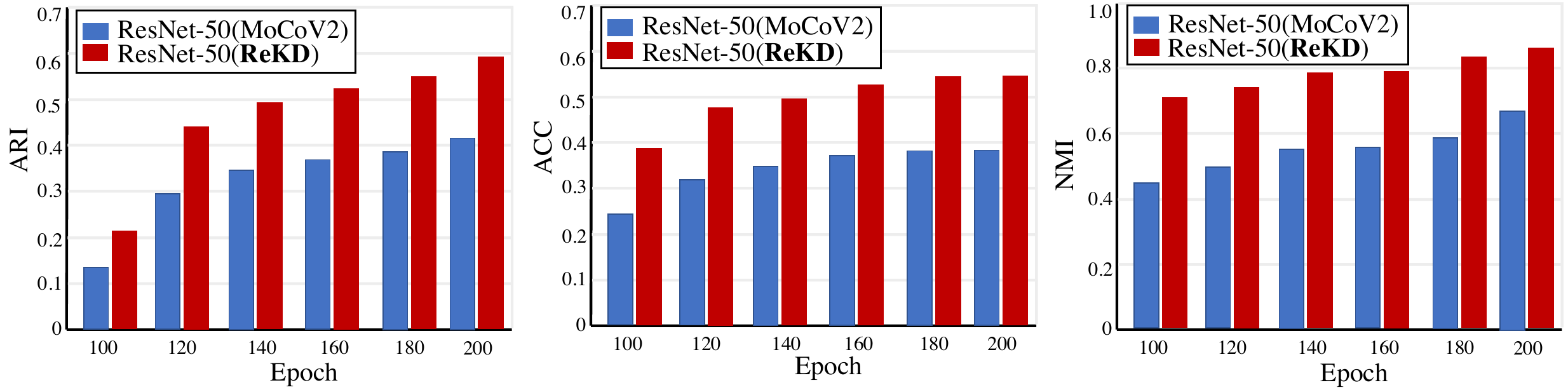}
\vspace{-3.5mm}
\caption{Adjusted rand index (ARI), normalized mutual information (NMI) and accuracy (ACC) calculated based on the prototype label inferred by ReKD and the ground-truth label from the ImageNet training data.}
\label{fig:ARI_NMI_ACC}
\vspace{-4.5mm}
\end{figure*}

\begin{figure}[ht]
\centering
\includegraphics[width=0.40\textwidth]{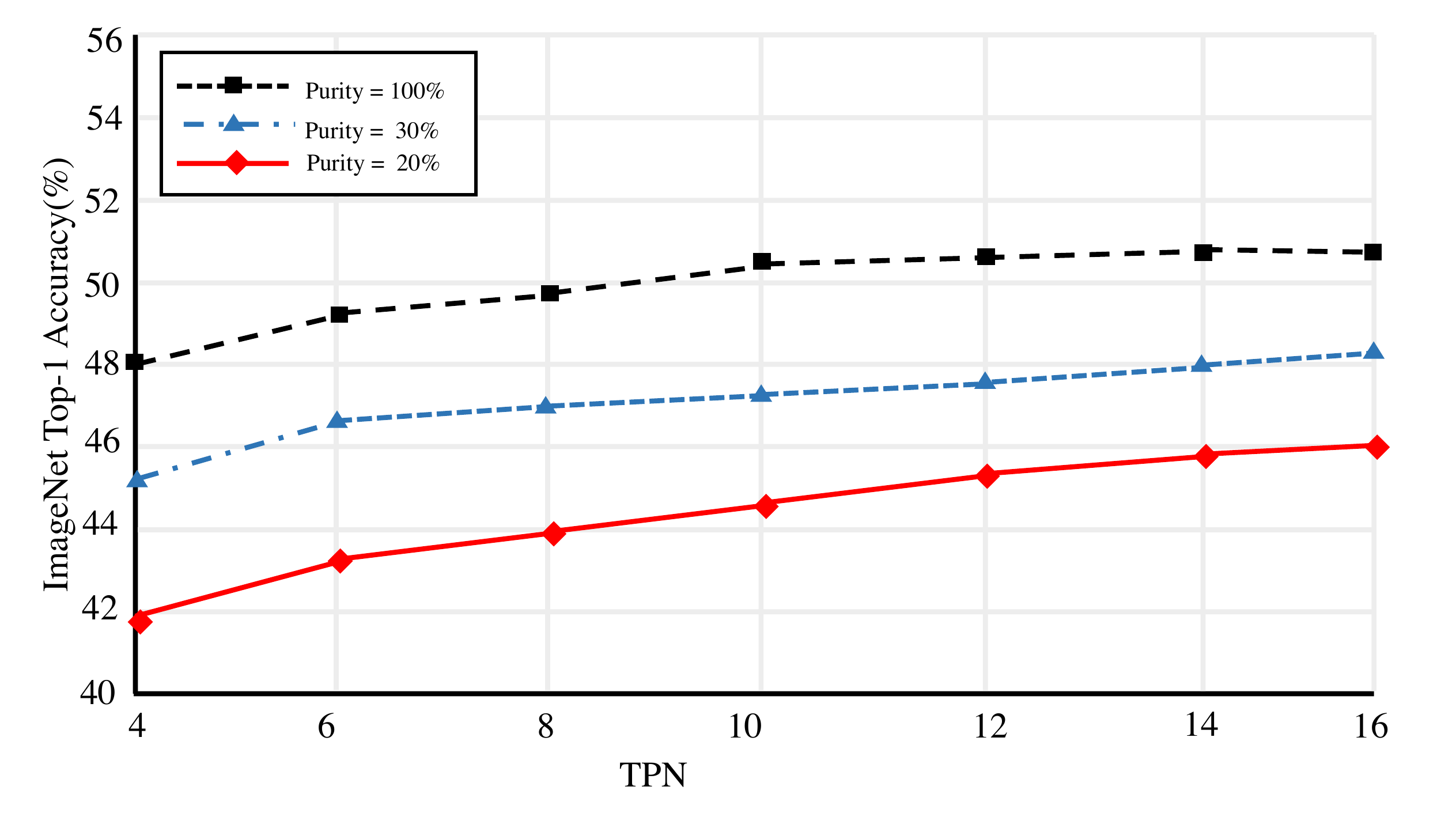}
\vspace{-3.5mm}
\caption{The relation among purity, true positive number and ImageNet top-1 accuracy(\%) under the Supervised MoCo experiment.}
\label{fig:tpn-purity}
\vspace{-4.5mm}
\end{figure}

\begin{figure*}[t]
\centering
\includegraphics[width=0.75\textwidth]{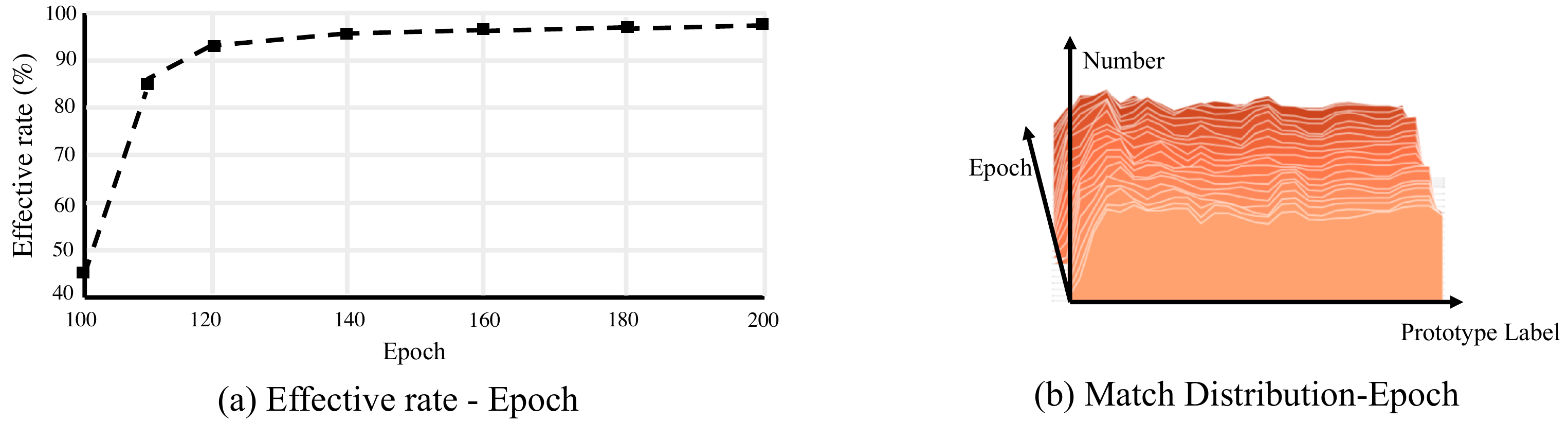}
\vspace{-2.0mm}
\caption{(a) \textnormal{\texttt{Effective Rate}} during training. (b) \textnormal{\texttt{Matching Distribution}} for assigned prototype label during the whole training.}
\label{fig:prototype_numeric}
\vspace{-4.5mm}
\end{figure*}

\section{Performance of Relation Knowledge.}
The \textit{relation knowledge} is used to indicate the semantic positives and negatives as supervision in ReKD. To analyze the performance of \textit{relation knowledge} statistically, we use \textnormal{\texttt{Positive Number}} (\textnormal{\texttt{PN}}) to denote the number of predictive positives mined for an \textit{anchor}. However, not all the positives are correct in the predictive positives. We thus denote the true positive number and the rate of true positives in all positives as \textnormal{\texttt{True Positive Number}}  (\textnormal{\texttt{TPN}}) and \textnormal{\texttt{Purity}}. As illustrated in Fig.~\ref{fig:tpn_purity} (a), we observe that AlexNet ({\textcolor[RGB]{30,144,255}{blue}}) has lower \textnormal{\texttt{TPN}} than ResNet-50 ({\textcolor[RGB]{50,205,50}{green}}), which means the ResNet-50 as the teacher has superior capability in mining accurate semantic positives. We can also observe that our ReKD ({\textcolor[RGB]{255,140,0}{orange}}) performs better than the baseline (MoCov2) under the same backbone in mining the positive. From Fig.~\ref{fig:tpn_purity} (a), we can see the \textnormal{\texttt{Purity}} has the same trend for three models. From the analysis of \textnormal{\texttt{Purity}} and \textnormal{\texttt{TPN}}, we conjecture that the \textit{relation knowledge} in ReKD transfers a superior semantic knowledge of mining the semantic similar positives to student, which significantly boosts the student's performance. Besides, the \textit{relation knowledge} is captured based on the prototype, where each \textit{candidate} can be assigned to a prototype as its label. Hence, we can estimate the normalized mutual information (NMI) between the assigned prototype and the ground-truth class label. We compare the NMI of our ReKD and that of MoCov2. As shown in Fig.~\ref{fig:tpn_purity} (b), ReKD have a larger NMI with the class labels due to the clustering effect from semantic prototypes, which indicates the relation knowledge retrieved from them is also semantically meaningful.

\section{Analysis From Supervised MoCo}\label{appendix:supmoco}
We provide an experiment on the SupMoCo (a.k.a Supervised MoCov2) to show the relation of \textnormal{\texttt{TPN}}, \textnormal{\texttt{Purity}} and ImageNet Acc. The SupMoCo is implemented based on MoCov2, where the label is given for relation contrastive loss to guide the positives and negatives. In the experiments, we can control the \textnormal{\texttt{Purity}} by replacing the ground-truth label with random label and choose the number of positives. We conduct the training of SupMoCo under different \textnormal{\texttt{Purity}}, \textnormal{\texttt{TPN}} levels. As can be seen in Fig.~\ref{fig:tpn-purity}, we can speculate that at different \textnormal{\texttt{Purity}} level, the more true positives exposed to SupMoCo, the better ImageNet Acc. can the model achieve. This conclusion also works for our ReKD which relies on the accuracy of positives.

\section{Analysis of Performance for Prototypes}\label{appendix:prototypeanalysis} In our method, prototypes are the medium to build connection between the \textit{anchor} and \textit{candidate}. Hence, it is necessary to analyze the prototypes. We discuss the prototype from several perspectives to prove its validity.
\paragraph{Effective Rate.} The \textit{candidate} set stores the embeddings with their prototype assignment, only the embedding whose maximum similarity is greater than the threshold will be assigned to the corresponding prototype, which we called effective embedding. The embedding that does not match any prototypes will be assigned as prototype \textnormal{\texttt{-1}} (a.k.a outlier), which is termed as ineffective embedding. We expect each embedding can be assigned to a prototype with high similarity. Thus, we define an \textnormal{\texttt{Effective Rate}} to indicate the rate of effective embedding in the \textit{candidate} set. As shown in Fig.~\ref{fig:prototype_numeric}(a), we observe that \textnormal{\texttt{Effective Rate}} gradually rises and tends to be \textnormal{\texttt{1}}, which means more samples are becoming effective and more compact with cluster centroids during the training.

\paragraph{Matching Distribution.} We expect the prototype-based clustering can form a good clustering distribution when matching the embeddings. A good clustering result needs to separate the samples equally into each cluster, especially for the dataset like ImageNet. Hence, we define a \textnormal{\texttt{Matching Distribution}} to denote all the embeddings' prototype assignment (cluster) distribution. Fig.~\ref{fig:prototype_numeric}(b) shows that matching distribution is uniform in all epochs, which means the clusters do not collapse. All the samples are well-distributed into each cluster.

\section{Analysis of Performance for Prototype Assignment}\label{appendix:cluster}
In order to evaluate the quality of the prototype assignment formed based on prototypes, we compute the Adjusted Rand Index (ARI), Normalized Mutual Information (NMI) and Accuracy (ACC) based on the prototype label and ground-truth label in the whole training. ARI evaluates the prediction label as a series of decisions and measures the prediction quality regarding how many positive/negative sample pairs are correctly assigned to the same/different prototypes. NMI measures the normalized mutual dependence between prediction labels and ground-truth labels. ACC is computed based on assigning each clustering with the dominant labels and use the average correct classification rate as the final result. In Fig.~\ref{fig:ARI_NMI_ACC}, we show ARI, NMI and ACC obtained by the teacher (ResNet-50) in ReKD in the whole training with the prototypes (clusters) number of $1000$. Note that, the sample number is $65536$ in computation.

\section{Pseudo Implementations}\label{appendix:pseudo_implementations}
We provide pseudo-code of the ReKD in PyTorch style.

\begin{algorithm}[t]
\caption{Pseudocode of ReKD in PyTorch style.}
\label{alg:code3}
\algcomment{\fontsize{7.2pt}{0em}\selectfont \texttt{mm}: matrix multiplication; \texttt{eq}: element-wise equality.
}
\definecolor{codeblue}{rgb}{0.25,0.5,0.5}
\lstset{
  backgroundcolor=\color{white},
  basicstyle=\fontsize{6.2pt}{6.2pt}\ttfamily\selectfont,
  columns=fullflexible,
  captionpos=b,
  commentstyle=\fontsize{7.2pt}{7.2pt}\color{codeblue},
  keywordstyle=\fontsize{7.2pt}{7.2pt},
}
\begin{lstlisting}[language=python]
# f_q, f_k: teacher encoder networks for anchor and candidate
# g_q, g_k: student encoder networks for anchor and candidate
# f_que: teacher's feature queue of L candidates
# g_que: student's feature queue of L candidates
# pidx_que: feature's prototype index queue
# m: momentum, t: temperature, c_thresh: confident threshold

protobank = InitProto(f_que) # initialize prototype bank KxC
f_k.params = f_q.params
g_k.params = g_q.params

for x in loader:
   # teacher
   logits_tea, k_tea = model_forward(x, f_q, f_K, f_que)
   k_sim = mm(k_tea, protobank.T) # NxK
   k_prob, k_idx = max(cdd_sim, dim=1)
   
   # similar mask: NxL
   sim_mask = eq(
      k_idx.view(N,-1).repeat(1,L),
      pidx_que.view(-1,L).repeat(1,L))

   # confident mask: NxL
   conf_prob = (k_prob >= c_thresh).bool()
   conf_mask = conf_prob.view(N,-1).repeat(1,L)

   # positive mask: NxL
   labels = (sim_mask & conf_mask).long()

   # relation contrastive loss, Eq.(4)
   tea_loss = RelCon(logits_tea/t, labels)
   
   # SGD update: anchor network
   tea_loss.backward()
   update(f_q.params)
   
   # momentum update: candidate network
   f_k.params = m*f_k.params+(1-m)*f_q.params

   # update feature queue and index queue
   enqueue_dequeue(f_que, pidx_que, k_tea, k_idx)

   # update prototype with features
   update_protobank(k_tea, k_idx, k_prob)

   # student
   logits_stu, k_stu = model_forward(g_q, g_k, g_que)

   # relation contrastive loss, Eq.(4), using teacher's label
   stu_loss = RelCon(logits_stu/t, labels)

   # SGD update: anchor network
   stu_loss.backward()
   update(g_q.params)
   
   # momentum update: candidate network
   g_k.params = m*g_k.params+(1-m)*g_q.params

   # update feature queue
   enqueue_and_dequeue(g_que, k_stu)
  
def model_forward(x, m_q, m_k, m_que):
   q = m_q.forward(aug(x)) # anchors: NxC
   k = m_k.forward(aug(x)) # candidates: NxC
   logits = mm(q.view(N,C), m_que.view(C,L)) # NxL
   return logits, k

# logits and labels with shape: NxL
def RelCon(logits, labels):
   neg = ((1 - labels) * torch.exp(logits)).sum(1)
   pos = (labels * torch.exp(-logits)).sum(1)
   loss = torch.log(1 + neg * pos)
   return torch.mean(loss)
\end{lstlisting}
\end{algorithm}

\section{Visualization of Retrieval based on Prototype}\label{appendix:prototype_retrieval}
\begin{figure*}[htb]
	\centering
	\includegraphics[width=0.8\textwidth]{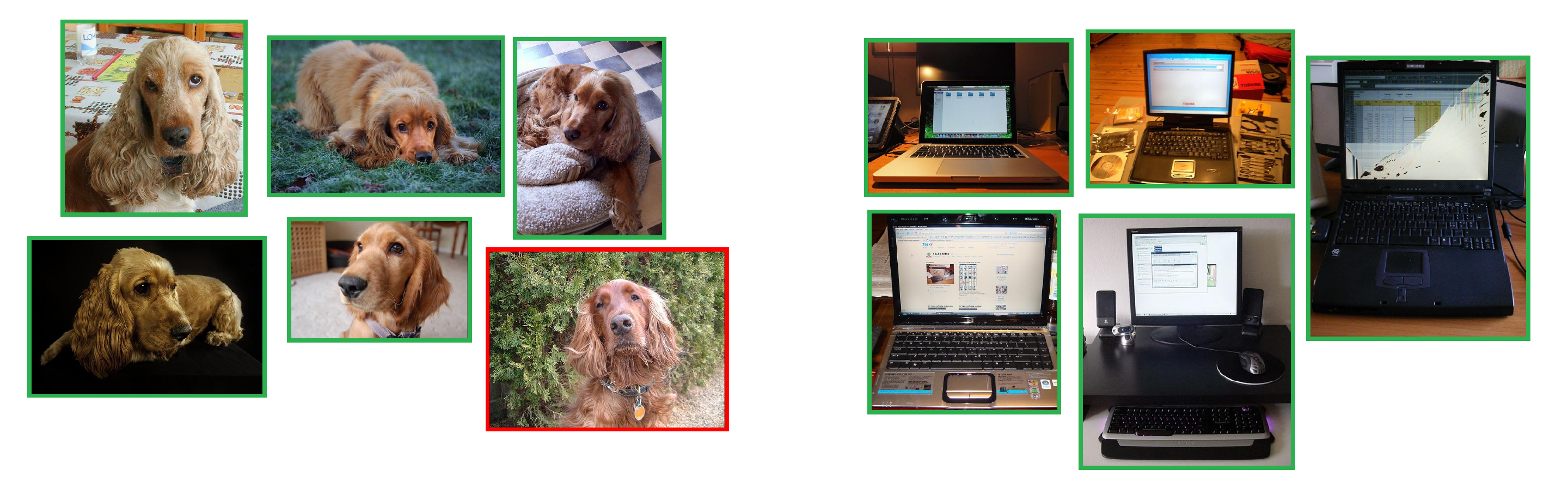} \\
	\vspace{.10in} 
	\includegraphics[width=0.8\textwidth]{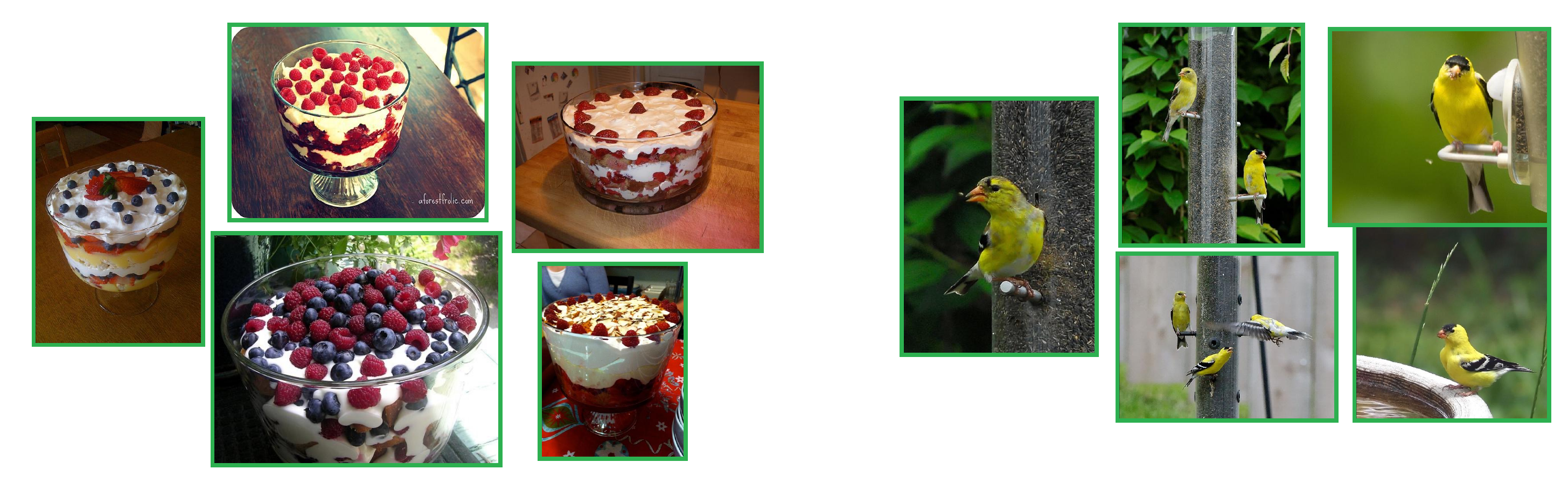}
	\vspace{.10in} 
	\includegraphics[width=0.8\textwidth]{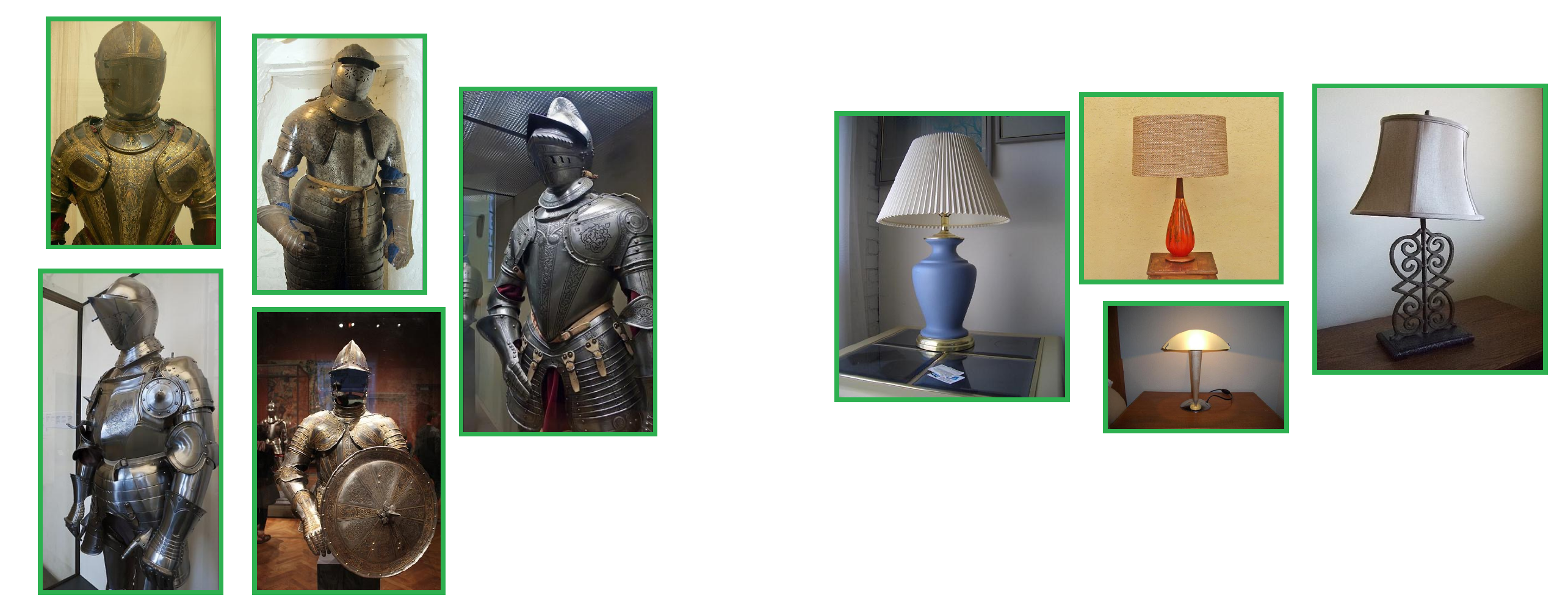}
	\vspace{.10in} 
	\includegraphics[width=0.8\textwidth]{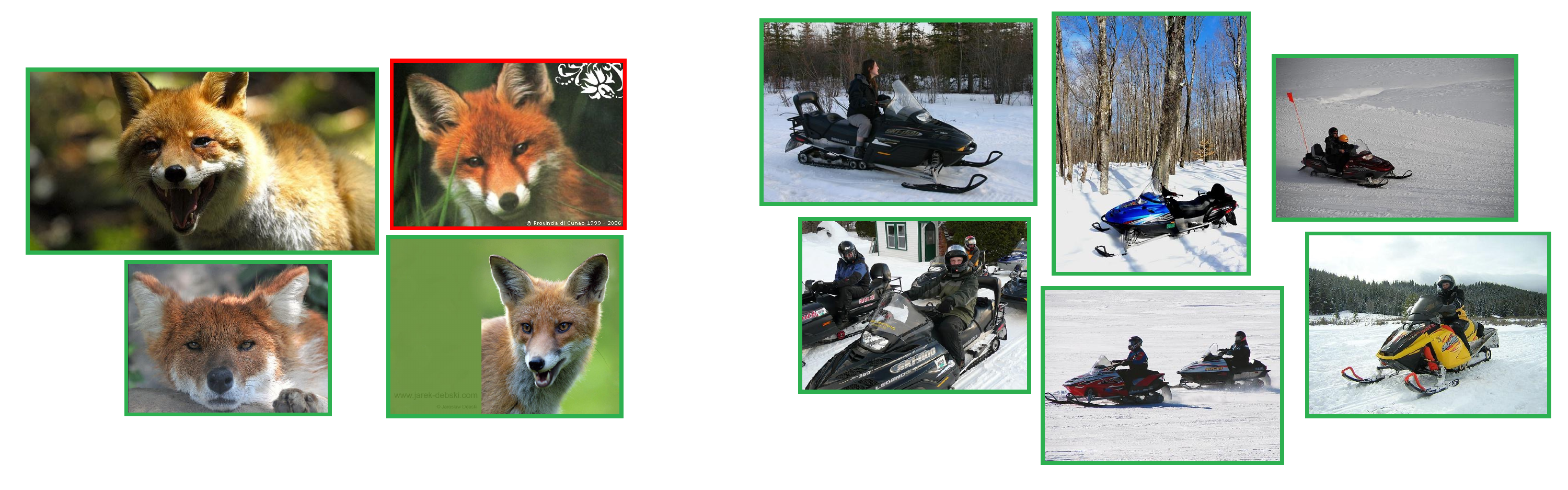}
	\caption{Visualization of retrieval based on prototypes generated by ReKD. The images in each clusters is retrieved by the same prototype based on cosine similarity. {\textcolor[RGB]{10,123,58}{Green}} boarder marks the same ground-truth label. {\textcolor[RGB]{255,0,0}{Red}} boarder marks the other label.}
	\label{fig:prototype_retrieval}
\vspace{-3.5mm}
\end{figure*}

In Fig.~\ref{fig:prototype_retrieval}, we show ImageNet training images that are retrieved by the prototypes. We see that different prototypes can learn the semantic information in the image and cluster the same semantic images together even in unsupervised learning.

\end{document}